\documentclass{llncs}
\usepackage{caption}
\usepackage{subcaption}
\usepackage{graphics} 
\usepackage{epsfig} 
\usepackage{mathptmx} 
\usepackage{times} 
\usepackage{amsmath} 
\usepackage{amssymb}  
\usepackage{tabularray}
\usepackage[table]{xcolor}
\definecolor{cvprblue}{rgb}{0.21,0.49,0.74}
\usepackage[pagebackref,breaklinks,
            colorlinks = true,
            linkcolor = cvprblue,
            urlcolor  = cvprblue,
            citecolor = cvprblue,
            anchorcolor = cvprblue]{hyperref}
\usepackage{calc}
\usepackage{cite}
\usepackage{booktabs}
\usepackage{multirow}

\usepackage{makeidx}  
\usepackage{etoolbox} 
\makeatletter
\patchcmd{\ps@headings}{\rlap{\thepage}}{}{}{}
\patchcmd{\ps@headings}{\llap{\thepage}}{}{}{}
\makeatother
\begin{document}
%
%

%
\mainmatter              
\title{Cognitive TransFuser: Semantics-guided Transformer-based Sensor Fusion \\for Improved Waypoint Prediction}
\titlerunning{Hamiltonian Mechanics}  
%
\author{Hwan-Soo Choi\inst{1}$^{,\blacklozenge}$, Jongoh Jeong\inst{3}$^{,\blacklozenge}$,
Young Hoo Cho\inst{2}, \\ Kuk-Jin Yoon\inst{3}$^{,\dagger}$, and Jong-Hwan Kim\inst{1,2}$^{,\dagger}$}
\authorrunning{Choi et al.} 
%
\tocauthor{Hwan-Soo Choi, Jongoh Jeong, Young Hoo Cho. Kuk-Jin Yoon and Jong-Hwan Kim}
\institute{Robot Intelligence Technology Lab., School of Engineering, \\ KAIST,  
Daejeon, Republic of Korea \\
\email{\{hschoi, johkim\}@rit.kaist.ac.kr} \\
\and
dSPECTER, Seoul, Republic of Korea \\
\email{\{yhcho, jhkim\}@dspecter.com} \\
\and
Visual Intelligence Lab., \\Robotics Program and Dept. of Mechanical Engineering, \\ KAIST, Daejeon, Republic of Korea \\
\email{\{jeong2, kjyoon\}@kaist.ac.kr}
\thanks{Code available at \href{https://github.com/Hwansoo-Choi/Cognitive-Transfuser}{https://github.com/Hwansoo-Choi/Cognitive-Transfuser} and video available at \href{https://shorturl.at/ahsyN}{shorturl.at/ahsyN}. 
$^{\blacklozenge}$Equal contribution (alphabetical order), $^{\dagger}$Co-corresponding authors.}
}

\maketitle

\begin{abstract}
Sensor fusion approaches for intelligent self-driving agents remain key to driving scene understanding given visual global contexts acquired from input sensors. Specifically, for the local waypoint prediction task, single-modality networks are still limited by strong dependency on the sensitivity of the input sensor, and thus recent works therefore  promote the use of multiple sensors in fusion in feature level in practice. While it is well known that multiple data modalities encourage mutual contextual exchange, it requires global 3D scene understanding in real-time with minimal computation upon deployment to practical driving scenarios, thereby placing greater significance on the training strategy given a limited number of practically usable sensors. In this light, we exploit carefully selected auxiliary tasks that are highly correlated with the target task of interest (e.g., traffic light recognition and semantic segmentation) by fusing auxiliary task features and also using auxiliary heads for waypoint prediction based on imitation learning. Our RGB-LIDAR-based multi-task feature fusion network, coined \textit{Cognitive TransFuser}, augments and exceeds the baseline network by a significant margin for safer and more complete road navigation in the CARLA simulator. We validate the proposed network on the Town05 Short and Town05 Long Benchmark through extensive experiments, achieving up to 44.2 FPS real-time inference time.
\end{abstract}

\section{Introduction}
Current advances in deep learning methods have paved the way for the maximum use of input sensors and various sensor fusion methods for integration of multiple sensors. In particular, for self-driven visual perception tasks that are crucial for road navigation (\textit{e.g.,} waypoint prediction, traffic light recognition, semantic segmentation, and lane detection), the type, quality, and quantity of input sensors play a significant role in increasing the recognition capacity of intelligent self-driving agents on the road. While early approaches to driving scene understanding mainly adopted 2D RGB camera as input, recent trends are geared towards efficient use of multiple synchronized sensors simultaneously to fully capture and reason in a global context and scope. 

Cost-effective RGB cameras have long demonstrated its capacity to capture sufficient coarse and fine details from 2D images for various visual driving perception tasks~\cite{bisenet, stdc, contextnet, rodsnet}. Similarly, LiDAR sensor-only methods have also shown the adequacy of LiDAR as a standalone data modality~\cite{LBC, codevilla2018end, CILRS, filos2020can, zhao2019sam}. While these learning-based computer vision algorithms with a single sensor modality have long been the conventional approach for their low resource costs and memory upon practical deployment of real-time systems, recent developments in hardware such as LiDAR sensor have led to reduced costs, thereby promoting the use of multiple modalities in fusion. Moreover, the emergence of the Transformer architecture using the attention mechanism~\cite{vaswani2017attention} led to a paradigm shift in deep learning, whereby numerous works such as \cite{chen2017brain, li2020end} have thus followed its traces to augment the network prediction capacities that depend on single- or multi-sensor inputs. In this light, \cite{transfuser} proposed the TransFuser network that efficiently fuses RGB camera and LiDAR data using transformers for imitation learning-based local waypoint prediction in the CARLA simulator~\cite{carla}, one of the important and final tasks in autonomous driving that has the most direct impact on occupant safety. However, due to its end-to-end nature designed for a single task only, it is inherently limited under certain difficult circumstances including driving through intersections. 

To mitigate this issue, we propose a waypoint prediction network guided by and cognizant of surrounding objects (\textit{e.g.,} traffic lights and road/non-road object pixels) called \textit{Cognitive TransFuser} in order to augment the RGB-LiDAR network~\cite{transfuser} to overcome shortcomings by exploiting complementary networks, each of which is trained for correlated auxiliary tasks. While many other multi-task networks that exploit global contextual exchange across task-specific features during training incorporate common and similar tasks such as semantic segmentation, we carefully design and incorporate semantically relevant traffic light recognition and semantic segmentation task modules for waypoint prediction, which are strictly visible from the ego vehicle and significantly affect the vehicle in road navigation.

\noindent
We highlight our contributions as follows:
\begin{itemize}
    \item We analyze different ways of fusing auxiliary tasks that are semantically meaningful for our target task and incorporate them in an efficient and modular approach, validated through exhaustive experiments.
    \item We demonstrate and validate the effectiveness of our multitask feature fusion method using complementary modules to improve the real-time imitation learning-based waypoint prediction in difficult circumstances during navigation (\textit{e.g.,} at intersections).
    \item Our proposed network outperforms the RGB-LiDAR-based baseline on driving score and route completion metrics by significant margins in complex adversarial urban driving scenarios in the CARLA simulator.
\end{itemize}

\section{Related Work}
\subsection{Sensor Fusion for Dynamic Scene Navigation}
        A number of recent works on autonomous driving have demonstrated effectiveness of using multiple data modalities to complement the conventional RGB camera-only approach~\cite{xiao2020multimodal, sobh2018end, liang2019multi, liang2018deep, munir2022multi}. As visual perception for self-driving agents requires understanding of 3D scenes in a global scope for safety and smooth navigation, adequate use of multiple data modalities is becoming more preferred for more stable and better scene understanding. In particular, for vision-based navigation control, \cite{xiao2020multimodal} fuses RGB and depth features in the early, mid, and late stages of the network for end-to-end driving scene sensing tasks, demonstrating the superiority of fusing RGB-D in the early stage based on a conditional imitation learning scheme. On the other hand, \cite{sobh2018end} suggests a late fusion method combining RGB, semantic segmentation, Bird's-Eye-View (BEV) LiDAR and Polar Grid Mapping (PGM) LiDAR features, whereby abstraction of segmentation feature maps along with PGM LiDAR features yields significantly better results than using raw RGB alone in navigational control. In this approach, we explore sensor fusion strategies at different stages of the network in this work.
        
        In a similar 3D object detection task, \cite{liang2019multi} explores direct exchange of intermediate feature maps of RGB image and BEV LiDAR point cloud, while \cite{ku2018joint} proposes to fuse front-view RGB and BEV LiDAR after a convolutional layer to predict k 3D region proposals which are fused back with each modality to finally detect 3D objects. \cite{zhao20193d, wang2019frustum}, on the other hand, fuse detected 2D objects from the front view RGB into the frustum LiDAR point cloud in data level to detect 3D objects.
    
    \subsection{Vision-based Visuomotor Control}
        With the emergence of open-source high-fidelity CARLA simulator~\cite{carla}, large-scale simulated driving has become available. Many recent works follow this line of research to introduce an imitation learning (IL)-based visuomotor control task guided by an expert algorithm available in the simulator~\cite{codevilla2018end, xiao2020multimodal, sobh2018end, CILRS, LBC, NEAT}. The objective of this behavioral cloning (BC)-based controller is to mimic the expert actions given a set of observations during navigation. While these methods focus on maneuvering the ego-vehicle parameters, \cite{transfuser} specifically aims to predict the PID control parameters for local waypoints given the expert goal locations determined by the expert A-star algorithm~\cite{hart1968formal}. Similarly to \cite{chen2017brain, li2020end}, \cite{transfuser} incorporates the attention mechanism for multi-sensor fusion for comprehensive and safe navigation in adverse scenarios with four transformer blocks corresponding to the feature embedding dimension at each resolution. we follow the traces of this work to improve the network by also attending to semantically meaningful auxiliary tasks.

\section{Method}

\begin{figure*}[!t]
    \centering
    \includegraphics[width=\textwidth]{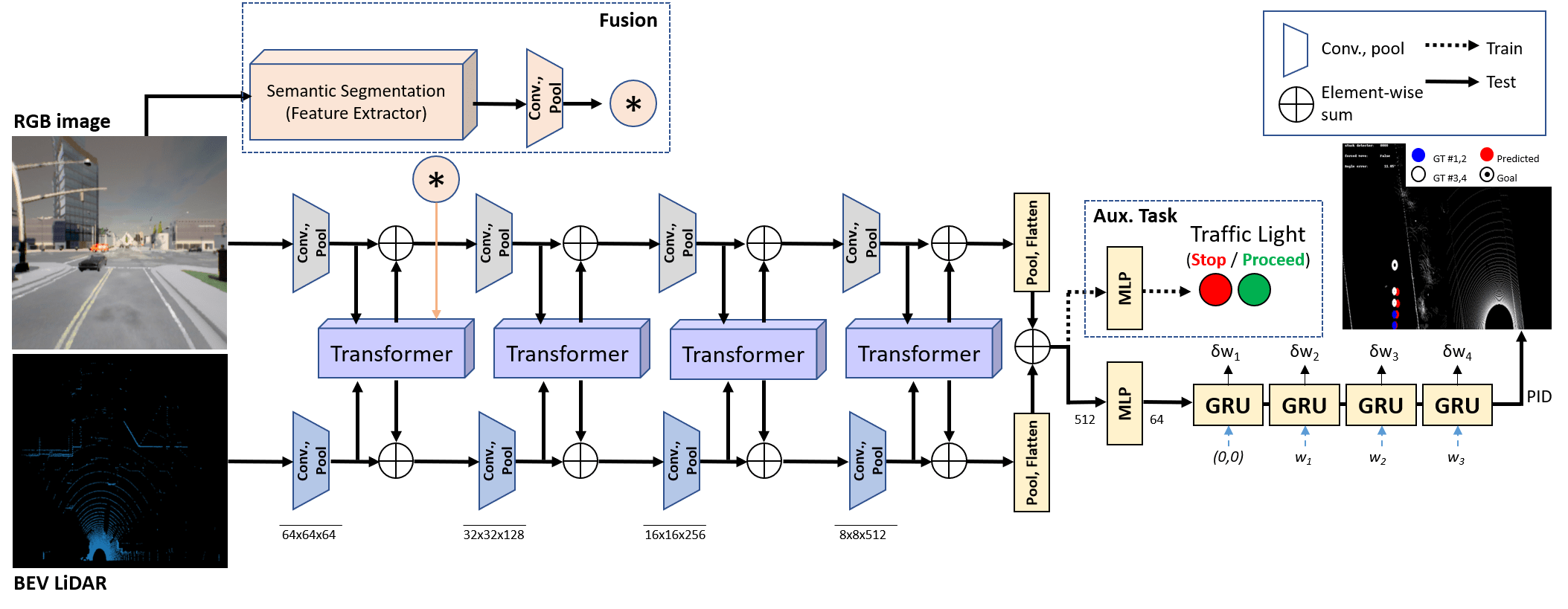}
    \caption{\textbf{Overview of \textit{Cognitive TransFuser}}. Given the front RGB image and the BEV LiDAR data, RGB is encoded by ResNet-34 backbone blocks in gray and LiDAR by ResNet-18 blocks in blue. We fuse the semantic segmentation feature map$^{\ast}$ into the first transformer fusion block, and sequentially extracted and merged features are used to predict the auxiliary traffic light classification label and local waypoints via a GRU sub-network. Please view in \textit{zoom} and \textit{color} for details.}
    \label{fig:overview}
\end{figure*}
    
\subsection{Problem Setting}
    Our task of interest deals with point-to-point local waypoint prediction for safe and complete road navigation in complex urban settings, following the previous works~\cite{CILRS,LBC,filos2020can,rhinehart2019precog,rhinehart2018deep}, whereby the ego vehicle is to complete a given route while handling other dynamic objects and obeying traffic rules simultaneously. We note that vehicle navigation in complex urban real-world scenarios for data collection is practically impossible, and thus we resort to following the expert based on IL, or BC, as in~\cite{transfuser}. As a form of supervised learning, we acquire ground truth data $\mathrm{D}$ of size $Z$ by running the expert policy $\pi$ in the environment, as in Eq.~\ref{eq:data_expert}:
    
    \begin{equation}
        \mathrm{D} = \{(\mathrm{X}_{i}, \mathrm{W}_{i})\}_{i=1}^{Z},
        \label{eq:data_expert}
    \end{equation}
    \noindent where high dimensional observations consist of RGB image $R_{t}$, Bird's Eye View (BEV) LiDAR point cloud $L_{t}$, and traffic light states $TL_{t}$ (stop or proceed), s.t. $\mathrm{X}_{t}=\{R_{t}, L_{t}, TL_{t}\}$, and the goal location $G$ of a global waypoint planner, position $P$ and velocity $V$ of the ego vehicle $\mathrm{W}_{t}=\{G_{t}, P_{t}, V_{t}\}$, respectively at time $t$. The waypoints are two-dimensional in BEV space, \textit{i.e.,} $\mathrm{P}=\{\mathbf{p}_{t} = (x_{t},y_{t})\}_{t=1}^{T}$. We then train the network in a supervised learning manner using the collected data $\mathrm{D}$ on the loss objective $\mathcal{L}$ as in Eq.~\ref{eq:objective}, followed by the manipulation of the PID controller for navigation control, consisting of steer, throttle, and brake:
    
    \begin{equation}
        \underset{\pi}{argmin}\; \mathbb{E}_{(\mathrm{X},\mathrm{W}) \sim \mathrm{D}}[\mathcal{L}(P, \pi(R,L,G,V))].
        \label{eq:objective}
    \end{equation}
    
\subsection{\textit{Cognitive TransFuser}}
\label{sec:cognitive_tranfuser}
\textit{Cognitive TransFuser} builds on top of the TransFuser~\cite{transfuser} architecture, which depends on a series of transformer blocks for the fusion of RGB and BEV LiDAR features propagated sequentially, as shown in Fig.~\ref{fig:overview}.. Front RGB image and BEV LiDAR data are fed into ResNet34 and ResNet18 encoder blocks, respectively, and the output of each encoder is fed as input to each transformer block for attentive fusion. The base network is then followed by a series of gated recurrent unit (GRU) modules, each of which predicts the positional change in the waypoint at the subsequent timestamp, \textit{i.e.,} $\{{\delta \mathbf{w}_{t}}\}_{t=1}^{T=4}$, where 4 is the default number of waypoints required by the inverse dynamics model for control. Given (0,0) fed into the first GRU as the initial position of the ego vehicle in BEV space, we obtain a sequence of waypoints $\mathbf{w}_{t} = \mathbf{w}_{t-1} + \delta \mathbf{w}_{t}$. 

To meet the real-time requirement for safety and smooth control of the ego vehicle in motion, we restrict the computation cost by pre-processing the inputs. We crop the front RGB (FoV=100$^{\circ}$) of 400$\times$300 to 256$\times$256 resolution, which also helps reduce the radial distortion at the edges, and the LiDAR point cloud collected over the frontal driving area of size 32m$\times$32m is converted into a 2-bin histogram on a 2D BEV grid with fixed 256$\times$256 resolution, following~\cite{sobh2018end, ku2018joint, liang2019multi}. The 2-bin histogram represents two discretized sets of points along the height dimension with respect to the ground plane, resulting in a two-channel pseudo-LiDAR image.

\begin{figure}[!t]
  \centering
  \includegraphics[width=\linewidth]{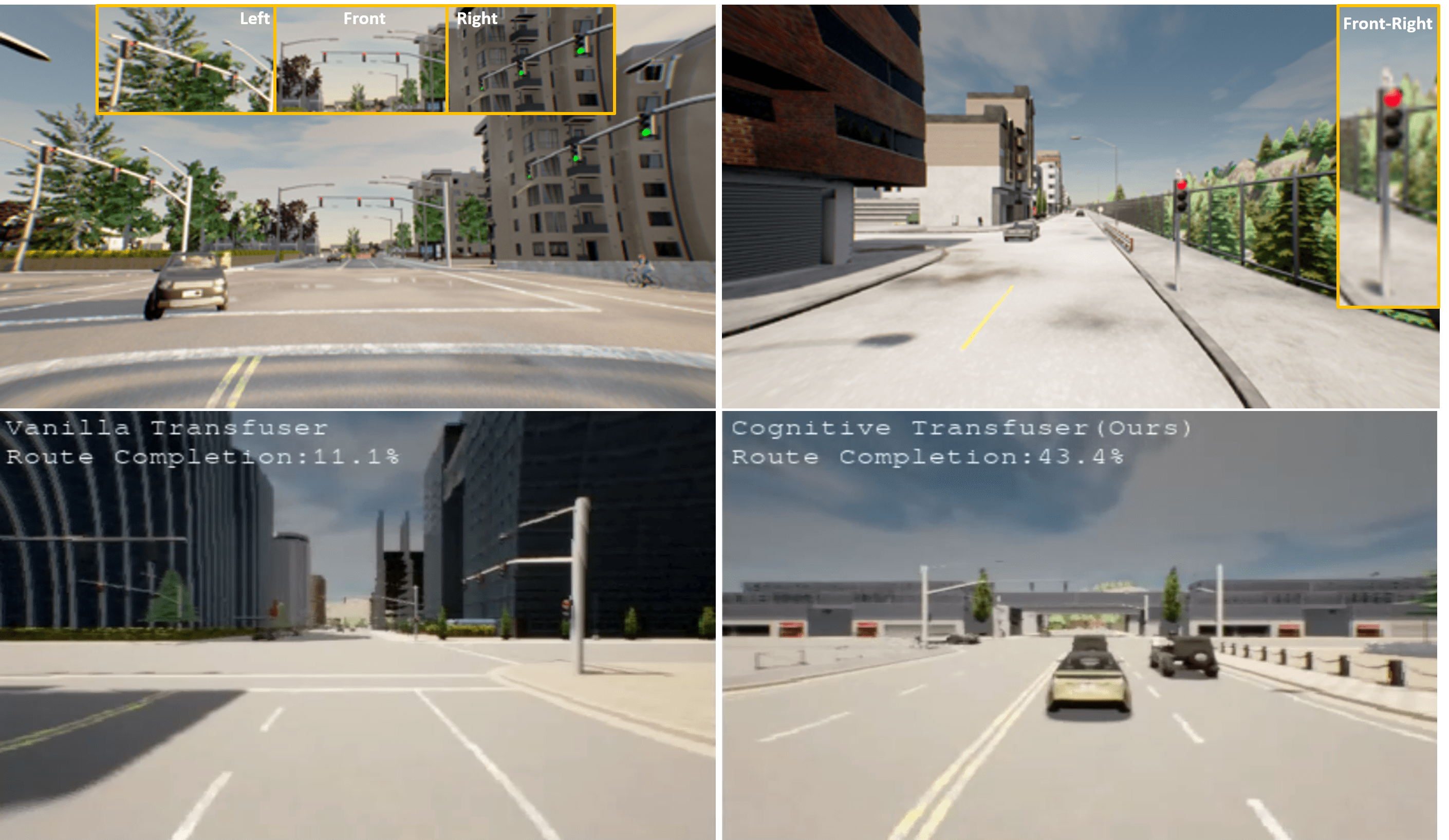}  
  \caption{\textbf{Top}: Sample observation instances of the ego vehicle at various traffic light signs (at a crossroad and along the sidewalk). \textbf{Bottom}: comparison of the driving scenes between the vanilla TransFuser and our \textit{Cognitive TransFuser} on the same time frame (see supplementary video).}
\label{fig:tf_sample}
\end{figure}

\begin{figure}[!t]
    \centering
    \includegraphics[width=\linewidth]{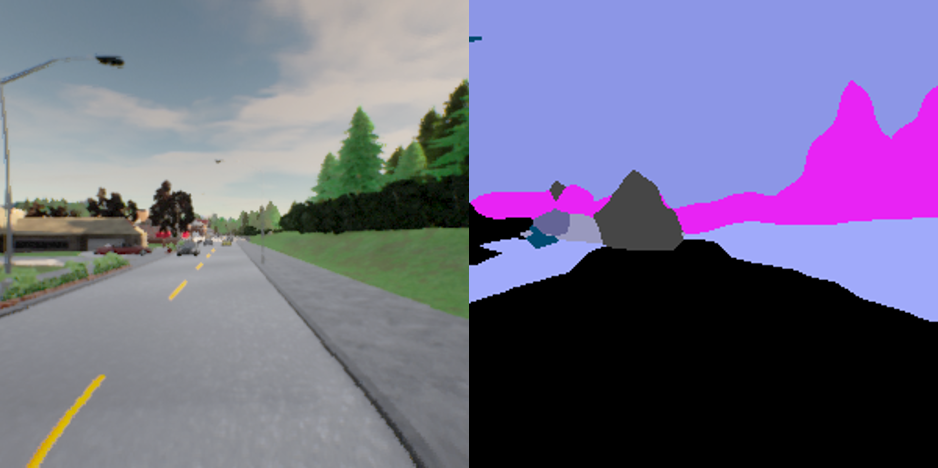}
    \caption{Sample semantic segmentation result using STDC-Seg50~\cite{stdc}.}
    \label{fig:rgb_seg}
\end{figure}

\noindent \textbf{Auxiliary task addition.} \quad
    We propose a simple yet effective modular multi-task feature fusion method to help the network to be more cognizant of correlated tasks to the target task of interest. To this end, we employ two auxiliary tasks that are semantically meaningful and highly correlated to the target waypoint prediction task: semantic segmentation and traffic light classification. As seen in Fig.~\ref{fig:tf_sample} and~\ref{fig:rgb_seg}, multiple traffic lights located at the intersection (\textit{e.g.,} on left, front, right sides) are key to the next path of the ego vehicle, and the semantic features hold object-specific localized information to guide the prediction of waypoints. Inspired by cross-modal attention statistics across fusion blocks~\cite{transfuser} that highlight greater attention weights on image features in earlier fusion as opposed to higher weights on LiDAR features in late fusion, we seek to aid early fusion where image features are further reinforced by semantic features from a pre-trained segmentation model. Note that the semantic feature map, $S^{aux} \in \mathbb{R}^{H \times W \times C}$, is acquired before the final decoder after average pooling, 

    As traffic lights play a key role in determining the procession of the ego vehicle, we also demonstrate the effect of adding features from a pre-trained traffic light detection model, whose results are outlined in Sec.~\ref{sec:exp}. Deciding whether to stop or proceed is one of the final decisions in driving that requires a thorough understanding of the environment~\cite{trafficlight_jensen2016vision, trafficlight_philipsen2015traffic}. Since traffic lights (1) are very small relative to large static objects, (2) resemble other objects in color (\textit{e.g.,} red), and (3) edge information is often not clear or sharp enough to be recognized as its class, the feature embedding from a trained 2D traffic light detection model holds highly useful semantics that can complement raw RGB. We do not consider 3D auxiliary tasks due to resource constraints. Similarly, the traffic light feature map $S^{aux} \in \mathbb{R}^{H \times W \times C}$ is taken just before the final decoder, after average pooling. We denote \textit{early, late, all} fusion as fusion into the first, fourth and all four transformer blocks. We adopt these fusion methods following the insights in~\cite{xiao2020multimodal, sobh2018end}, while we did observe much worse performance with fusion at the second and third blocks.

    In addition to leveraging the task-specific features in fusing alongside 2D RGB and 3D LiDAR features, we also add a prediction head for the auxiliary task to further promote globally contextual multi-task learning scheme. With dimension of the final features before the prediction head as 1$\times$512, the additional auxiliary task head outputs the softmax probability for each class: Stop/Proceed for TL and segmentation map following the standard Cityscapes~\cite{cordts2016cityscapes} categories, after fully connected layers (1$\times$64) and ReLU.

\section{Experiments}
\label{sec:exp}

    \subsection{Dataset and Evaluation Metric}
    
    To evaluate our trained network, we regenerated our data set following the standard protocol of CARLA 0.9.10 and the data acquisition strategies practiced by~\cite{transfuser}. We ran a built-in expert policy $\pi$ in CARLA on our collected dataset obtained from 8 CARLA towns except Town05, which is held out for testing. In each of the driving scenarios in the training set, we set the vehicle to navigate through scenes with dynamic agents and weather conditions on a frame-by-frame basis. The 14 CARLA weather variations\footnote{https://carla.readthedocs.io/0.8.4/carla\_settings} during training are designed to ensure that the prediction can be run robustly under rapid weather changes. On the other hand, we fix the test condition as \textit{ClearNoon}. The acquired data modalities are RGB camera, LiDAR point cloud, and annotations include ground truth waypoint positions, semantic segmentation maps, and binary traffic light signs (train only) for auxiliary tasks. In Fig.~\ref{fig:tf_sample}, we demonstrate several type of instances where the ego vehicle observes various traffic light signs at various locations (\textit{e.g.,} at crossroads, along sidewalk) (see the supplementary video for navigation samples). In Fig.~\ref{fig:rgb_seg}, we also show a sample predicted semantic segmentation map from STDC-Seg50, which is used as an auxiliary module in our network.
    
    
    %
    %
    %
    We follow the evaluation metrics established by the CARLA Autonomous Driving Challenge\footnote{https://leaderboard.carla.org/challenge}: Driving Score (DS) and Route Completion (RC). RC score is the percentage of route distances completed by the agent averaged across $N$ routes, with the penalty 1(-\% off route distance) given if the agent drives outside the planned route lanes for a percentage of the route, as described in Eq.~\ref{eq:RC}:
    \begin{equation}
        \textbf{RC} = \frac{1}{N} \sum_{i}^{N} R_{i}.
        \label{eq:RC}
    \end{equation}
    

    \noindent
    DS is the averaged route completion score with the predefined infraction penalty multiplier $P_{i}=\{0.50, 0.60, 0.65, 0.7\}$ for pedestrians, vehicles and static layout and for red light violations, respectively (Eq.~\ref{eq:DS}):
    \begin{equation}
        \textbf{DS} = \frac{1}{N} \sum_{i}^{N} R_{i}P_{i}.
        \label{eq:DS}
    \end{equation}
            



\subsection{Implementation Details}

    \begin{table*}[!ht]
        \centering
        \setlength\tabcolsep{0.01pt} 

        \caption{\textbf{Road navigation results on the Town05 Benchmark}. \textit{TL} and \textit{SS} denote traffic light and semantic segmentation modules, respectively. The traffic light infraction score indicates the average number of red light violations as the ego vehicle navigates the routes. We focus on TL violations for TL fusion and added computations for SS fusion. Inference time is measured on a single Nvidia GTX 1080Ti and i7-9700K. The results are highlighted as the \textbf{best} and \underline{second best}.}
        \resizebox{\linewidth}{!}{%
        \begin{tabular}{lcccccc}
             \toprule
              & \multicolumn{2}{c}{\textbf{Town05 Short}} & \multicolumn{2}{c}{\textbf{Town05 Long}} & \textbf{Avg. \# Red TL Violations} $\downarrow$  & \textbf{Inference Time} (ms)\\
             \cmidrule{2-5}
              \multirow{1}{*}{\textbf{Method}} & DS $\uparrow$ & RC $\uparrow$ & DS $\uparrow$ & RC $\uparrow$ & (per distance traveled; Town05 Short) & (additional $\Delta$t) [FPS]\\
            \midrule
    
            CILRS~\cite{CILRS} & 7.47 & 13.40 & 3.68 & 7.19 & - & - \\
            LBC~\cite{LBC} & 30.97 & 55.01 & 7.05 & 32.09 & $>$11 ($>$0.140) & - \\
            TransFuser~\cite{transfuser} (baseline) & 54.52 & 78.41 & 33.15 & 56.36 & $>$8 ($>$0.102) & 16.5\\
            NEAT~\cite{NEAT} & 58.70 & 77.32 & 37.72 & 62.13 & - & - \\
            Roach~\cite{Roach} & 65.26 & 88.24 & 43.64 & 80.37 & - & - \\
            WOR~\cite{WOR} & 64.79 & 87.47 & 44.80 & 42.41 & -  & -\\
            
            \midrule \midrule
            
            \textbf{Baseline + Feature fusion} \\
            Early fusion (TL) & \textbf{70.76} & \underline{78.81} & 35.38 & 77.70 & 1.7 (0.022) & -\\
            Late fusion (TL) & 54.32 & 66.73 & 32.59 & 72.62 & 1.3 (0.019) & - \\
            \textit{Early fusion (SS)} & \underline{69.45} & \textbf{87.46} & \underline{40.87} & \textbf{89.63} & - & 22.6 (+6.1) [44.2] \\ 
            All fusion (SS) & 56.79 & 67.85 & \textbf{48.87} & \underline{87.15} & - & 27.4 (+10.9) [36.5] \\ 
            \midrule
            
            \textbf{Baseline + Aux. task head (train only)} \\
            \textit{Aux. Head (TL)} & \textbf{79.34} & \textbf{90.82} & \textbf{53.21} & \underline{95.76} & 1.7 (0.019) & 16.5 (-) [60.6] \\
            Aux. Head (SS) & 63.37 & 73.09 & 45.15 & 86.43 & $>$8 ($>$0.102) & 16.5 (-) [60.6] \\
            Aux. Head (TL \& SS) & \underline{78.87} & \underline{89.12} & \underline{50.12} & \textbf{96.17} & 1.7 (0.019) & 16.5 (-) [60.6]\\
            \midrule \midrule
            
            \rowcolor{gray!30} \textbf{Early fusion (SS) + Aux. Head (TL)} (\textit{Ours}) & \textbf{80.67} & \textbf{95.14} & \textbf{52.70} & \textbf{96.18} & 1.7 (0.019) & 22.6 (+6.1) [44.2] \\ 
            \bottomrule
        \end{tabular}
        }
        \label{tab:results}
    \end{table*}

    We use 2D RGB camera and 3D BEV-converted LiDAR point cloud as input modalities, which are, respectively, encoded by ImageNet-pretrained ResNet34 and raw ResNet18. For each fusion block, we use one transformer with 4 attention heads for each feature dimension $D_{f}$=$D_{q}$=$D_{k}$=$D_{v}$=$\{64,128,256,512\}$ as in Fig.~\ref{fig:overview}. We follow other configurations (\textit{e.g.,} multi-scale fusion, positional embedding, multiple attention layers) as in \cite{transfuser} for a fair comparison on the CARLA towns. 

    Unless otherwise stated, we performed all of our experiments on a single GTX 1080Ti and i7-9700K using PyTorch. To extract RGB and LiDAR features, we used ResNet34 and ResNet18, respectively, using an Adam optimizer with a learning rate of 1$\times$10$^{-4}$ and a batch size of 10. Each of the four transformer fusion blocks has output feature dimensions of 64, 128, 256, and 512. Note that for Aux. Head (TL \& SS) in Table~\ref{tab:results}, we empirically tuned the loss coefficients to $\lambda_{aux}^{tl}:\lambda_{aux}^{ss}:\lambda_{wp}=1:0.3:1$ for the best performance, as the TL guidance was more meaningful than the SS.  
    
    \noindent
    \textbf{Task-specific pre-training.} \quad
    Prior to fusing additional semantically meaningful information into the fusion blocks, we pre-trained external real-time-capable networks for each semantic segmentation and traffic light detection tasks. For the former task, we trained STDC-Seg50~\cite{stdc} on Cityscapes containing 2,975 annotated training images with SGD (momentum $\eta=0.9$) optimizer, weight decay of 5$\times$10$^{-4}$ and batch size of 48. For the semantic segmentation task, we use STDC-Seg whose output is shown in Fig.~\ref{fig:rgb_seg} to achieve the highest mean Intersection-over-Union (mIoU) score while minimally compromising the inference FPS. We augment data using color jittering, random horizontal flip, random crop to 1024$\times$ 512, and random scaling to range $[0.125, 1.5]$, yielding the test accuracy of 71.9 mean intersection-over-union (mIoU) at 250.4 FPS on a GTX 1080Ti. 
    For the latter, we trained YoloV3~\cite{yolov3} object detection model on LISA Traffic Light (TL) dataset~\cite{trafficlight_jensen2016vision, trafficlight_philipsen2015traffic} containing 43k frames with 113k traffic light annotations, consisting of six TL types: \textit{Stop, Proceed, Warning, WarningLeft, ProceedLeft, StopLeft} and bounding boxes in (x, y, width, height) format. We pre-trained the YoloV3 model for 300 epochs and optimized with the default loss objective using SGD ($\eta=0.937$) (learning rate = 0.01), and decayed weights using a cosine scheduler until 5$\times$10$^{-4}$, yielding mean average precision (mAP) of 0.919. Note that while we pre-trained with multiple TL labels, we particularly focus on learning whether to \textit{Stop} or \textit{Proceed} using our auxiliary TL head. The front image resolution was set to 426$\times$426 since 255$\times$255 fails to sufficiently detect traffic lights on YoloV3.

\subsection{Experimental Results}
    
    We compare our driving test and red light infraction scores, as well as inference times, between various fusion methods, in Table~\ref{tab:results}. Overall, we observed early fusion of SS features and using TL auxiliary task head outperformed other methods by significant margins in DS, RC and red light violation frequency. Compared to the baseline, \textit{Cognitive TransFuser} improves DS by 26.15\%p and RC by 16.73\%p in Town05 Short, DS by 19.55\%p and RC by 39.82 \%p in Town05 Long, and the average number of traffic light violations by at least 6.3. Although the inference time increased slightly, 44.2 FPS remains promising for real-time capability. Furthermore, a much lower red light violation rate than the baseline, as well as notable improvements in DS and RC, ensures safer and more complete navigation, respectively.

    \noindent\textbf{Auxiliary: traffic light classification.} \quad
    The baseline is mainly limited since it had a high red light violation rate at crossroads, and upon sudden stopping of the ego vehicle, traffic light violations could not be measured properly. Meanwhile, our method effectively leverages the TL head during training only, thereby lowering violations without increasing the inference time due to the addition of the TL head. We posit that late fusion of TL lowers TL violations but disturbs learning of direct information associated with waypoints, while early fusion aids in learning of spatial information. Meanwhile, the auxiliary TL classification task outperformed the fusion methods, as the learning is oriented by direct ground-truth guidance. Specifically, the ego vehicle should only attend to TL signs located in its front view as shown in Fig.~\ref{fig:tf_sample} and the TL classification label resolves the difficulty of assessing which TL sign holds stronger relation to subsequent waypoints at crossroads. Given TL label, the addition of semantic segmentation features at an earlier stage allows to capture more global context in combination with a TL head. Although the TL head is used only in training, we remark that it is highly dependent upon the availability of the GT label.

    \noindent\textbf{Auxiliary: semantic segmentation.} \quad
    The outperforming driving scores in Town05 in early SS fusion than all fusion insinuates the merit of incorporating semantic features early alongside RGB and LiDAR data for global contexts. Early fusion is also computationally advantageous in memory and time compared to all-fusion methods in inference. All SS fusion yields worse results due to the excessive SS task-specific guidance across all fusion blocks. When added as an auxiliary task head, the addition of SS did not provide better GT guidance than TL towards waypoint prediction since waypoint is more affected by the vehicle procession than mere location of pixel-level objects. However, in the combined early SS fusion and TL auxiliary head, their respective roles were maximized, even surpassing different architectures~\cite{NEAT, Roach, WOR}. 


\section{Conclusion}
In this paper, we presented \textit{Cognitive TransFuser} that leverages two complementary task modules for improved global contextual reasoning, specifically to aid the local waypoint learning based on imitation learning from the expert. We introduced an effective method to incorporate correlated tasks as features and multi-task settings in order to help guide the transformer-based sensor fusion network to better predict waypoints. Through extensive studies on fusion methods for RGB camera and LiDAR modalities, we verified that fusing semantic features in the early fusion block and training with the traffic light classification head additionally further aids the ego vehicle in safer and more complete road navigation. 

\noindent \textbf{Limitations and future work.} \quad
Although our proposed feature fusion method achieves both high performance and real-time applicable inference capability, we acknowledge that there exists limitations in which task features are useful for other target tasks and RGB and LiDAR sensor modalities. With other usable sensors such as RADAR, dynamic vision sensor, and thermal camera in mind, we hope to extend to multi-modal feature fusion with auxiliary task guidance in the future.  We hope our work sheds further light on simple yet effective sensor fusion approaches for real-time self-driving systems.
  
\noindent \textbf{Acknowledgements.} This work was supported by the Starting growth Technological R\&D Program (TIPS) (No. RS-2023-00261771) funded by the Ministry of SMEs and Startups (MSS, Korea) and by the Institute for Information \& Communications Technology Promotion (IITP) grant funded by the Korea government (MSIT) (No.2020-0-00440, Development of Artificial Intelligence Technology that Continuously Improves Itself As the Situation Changes in the Real World).
\bibliographystyle{unsrt} 
\bibliography{llncs.bib}







\end{document}